\documentclass{article}

\usepackage{microtype}
\usepackage{graphicx}
\usepackage{subfigure}
\usepackage{booktabs} 

\usepackage{hyperref}

\usepackage[accepted]{icml2024}

\usepackage{amsmath}
\usepackage{amssymb}
\usepackage{mathtools}
\usepackage{amsthm}

\usepackage[capitalize,noabbrev]{cleveref}

\theoremstyle{plain}
\newtheorem{theorem}{Theorem}[section]

\newtheorem{lemma}[theorem]{Lemma}

\theoremstyle{definition}

\theoremstyle{remark}

\usepackage[textsize=tiny]{todonotes}

\usepackage{macros}

\usepackage[textsize=tiny]{todonotes}

\newcommand{\I}[1]{\mathds{1} \! \left\{#1\right\}}
\newcommand{\realset}{\mathbb{R}}

\newcommand{\cA}{\mathcal{A}}

\newcommand{\ie}{\textit{i.e.}}

\icmltitlerunning{Efficient and Interpretable Bandit Algorithms}
\allowdisplaybreaks
\begin{document}

\twocolumn[
\icmltitle{Efficient and Interpretable Bandit Algorithms}




\icmlsetsymbol{out}{*}

\begin{icmlauthorlist}
\icmlauthor{Subhojyoti Mukherjee}{yyy}
\icmlauthor{Ruihao Zhu}{comp}
\icmlauthor{Branislav Kveton$^*$}{sch}
\end{icmlauthorlist}

\icmlaffiliation{yyy}{ECE Department, University of Wisconsin Madison, USA}
\icmlaffiliation{comp}{SC Johnson College of Business, Cornell University, USA}
\icmlaffiliation{sch}{AWS AI Labs, Amazon, USA}

\icmlcorrespondingauthor{Subhojyoti Mukherjee}{smukherjee27@wisc.edu}
\icmlcorrespondingauthor{Ruihao Zhu}{ruihao.zhu@cornell.edu}
\icmlcorrespondingauthor{Branislav Kveton}{bkveton@amazon.com}

\icmlkeywords{Machine Learning, ICML}

\vskip 0.3in
]



\printAffiliationsAndNotice{\icmlOutsideContribution} 

\begin{abstract}
Motivated by the importance of explainability in modern machine learning, we design bandit algorithms that are \emph{efficient} and \emph{interpretable}. A bandit algorithm is interpretable if it explores with the objective of reducing uncertainty in the unknown model parameter. To quantify the interpretability, we introduce a novel metric of \textit{model error}, 
which compares the rate reduction of the mean reward estimates to their actual means among all the plausible actions.
%
%
We propose \code, a bandit algorithm based on a \textbf{C}onstrained \textbf{O}ptimal \textbf{DE}sign, that is interpretable and maximally reduces the uncertainty. The key idea in \code is to explore among all plausible actions, determined by a statistical constraint, to achieve interpretability. We implement \code efficiently in both multi-armed and linear bandits and derive near-optimal regret bounds by leveraging the optimality criteria of the approximate optimal design. \code can be also viewed as removing phases in conventional phased elimination, which makes it more practical and general. We demonstrate the advantage of \code by numerical experiments on both synthetic and real-world problems. \code outperforms other state-of-the-art interpretable designs while matching the performance of popular but uninterpretable designs, such as upper confidence bound algorithms.
\end{abstract}

\section{Introduction}
A \emph{multi-armed bandit (MAB)} is a class of sequential decision-making problems that have gained significant attention in various fields, including machine learning, economics, and operations research, due to their broad applicability in real-world scenarios, such as clinical trials \citep{thompson1933likelihood, agrawal2012analysis}, online advertising \citep{li2010contextual, foster2021efficient}, and recommender systems \citep{li2016collaborative, eide2018deep, bendada2020carousel, mehrotra2020bandit}. In MAB, the goal is to maximize the cumulative reward by interacting with a set of actions under an initially unknown model parameter. Therefore, we have to balance exploration (learning the parameter) and exploitation (reward maximization). Two notable approaches that efficiently balance exploration and exploitation are \emph{upper confidence bound (UCB)} \citep{auer2002finite,abbasi2011improved} and \emph{Thompson sampling (TS)} \citep{agrawal2012analysis, agrawal2013thompson}. At a high level, UCB is a frequentist approach that uses the estimated mean reward of each action and its confidence interval to strike the right balance. TS is a Bayesian approach that maintains a posterior distribution of the model parameter, and samples actions accordingly. 

Perhaps not surprisingly, apart from statistical efficiency, another important factor in modern machine learning is \textit{interpretability} \citep{carvalho2019machine, molnar2020interpretable}. Interpretability can increase human trust in machine learning models for certain critical tasks, such as medical diagnoses. Therefore, it is critical to study interpretability in MABs \citep{huyuk2022inverse}. In the context of MABs, some of the most important quantities of interest are the mean reward of the optimal and near-optimal actions. For example, in clinical trials, besides the best treatments, one is also interested in the optimal and near-optimal treatment effects. Knowing these values can help to determine if one should proceed to the next phase of the trials and which treatments should be included. Moreover, as shown in \citep{garivier2018explore} (see Appendix C), when the optimal mean reward is known, one can achieve constant regret in MABs. Therefore, the near-optimal mean rewards of one MAB task can help to inform future decision-making in similar MABs (\textit{e.g.,} when some of these actions remain available). Following these ideas, we define interpretability as learning the mean rewards of near-optimal actions accurately. To further take reward maximization into account, we say that a bandit algorithm that satisfies the following as both \emph{efficient} and \emph{interpretable}: 
\begin{quote}
  \textit{A bandit algorithm is efficient and interpretable if it takes actions to attain low regret and learn the mean rewards of near-optimal actions accurately.}
\end{quote}

In this paper, we study the design of efficient and interpretable bandit algorithms. To this end, we make the following two observations:
\begin{itemize}
\item While both UCB and TS are efficient because they maximize the cumulative reward by exploring at a near-optimal rate, neither of them does a good job in learning the rewards of near-optimal actions (see the forthcoming \cref{sec:experiments});
\item  On the other hand, \emph{explore-then-commit (EtC)} algorithms \citep{langford08epochgreedy,garivier2016explore,jin2021double}) are interpretable. This is because they first explore to minimize model parameter uncertainty, which gives a good estimate of each action's reward, and then commit to the best empirical action. However, they are statistically sub-optimal due to non-adaptivity and may incur high regret.
\end{itemize}

As a partial remedy, several phased elimination algorithms have been proposed \citep{soare14bestarm, fiez2019sequential,katz2020empirical,wagenmaker2021experimental, mukherjee2022chernoff}. These algorithms run in phases and eliminate sub-optimal actions in between. Within each phase, they rely on an optimal design \citep{pukelsheim2006optimal, fedorov2010optimal} to find a distribution over actions that maximally explore the unknown parameter, to learn it. As a result, they are more efficient than EtC as they are more adaptive. In addition, they are interpretable because their exploration is guided by optimal designs over plausible, or near-optimal (given the collected data), actions. This helps to minimize the uncertainty of the mean reward estimates of the plausible actions.
However, these algorithms \begin{enumerate}
  \item Perform poorly empirically since no sub-optimal actions are removed in the middle of a phase, even if they are detected early within the phase; and
  \item More critically, cannot handle changing action sets (see Note 1 in Section 22.1 of \citet{lattimore2020bandit}), which is often the case in contextual bandits.
\end{enumerate}

We try to bridge the gap between efficiency and interpretability. Specifically, we make the following contributions:
\begin{itemize}
\item We propose \code, a bandit algorithm based on a \textbf{C}onstrained \textbf{O}ptimal \textbf{DE}sign, that is both efficient and interpretable. In each round, \code uses an approximate version of the $D$-optimal design to explore all plausible actions, determined by a statistical constraint that excludes sub-optimal actions. This leads to a statistically sound exploration guided by the optimal design that maximally reduces the uncertainty in the model parameter. 
Critically, \code avoids inefficient phases in phased elimination algorithms by replacing the elimination step with the constraints, and hence, can deal with time-varying action sets. 

\item Motivated by optimal designs \citep{pukelsheim2006optimal, fedorov2010optimal}, we propose the first metric for interpretability in the bandit setting. This metric estimates the reduction in model uncertainty. We quantify our metric by estimating the mean reward estimates of all the plausible actions to their actual means. The faster this reduction occurs, the better is the model estimate of the algorithm. We call it the \emph{model error}.



\item Next we instantiate \code in both multi-armed and linear bandits and show that it attains near-optimal regret. Unlike in conventional analysis, the regret of \code in each round depends on the confidence intervals of both the taken and optimal actions. We leverage the optimality criteria of the approximate optimal design problem to control the confidence interval of the optimal actions. 

\item Finally, we demonstrate empirically that \code outperforms other state-of-the-art interpretable bandit algorithms in linear bandits, in both synthetic problems and simulations based on real-world data. 
\end{itemize}

This paper is organized as follows. In \Cref{sec:model}, we introduce our notation and problem setting. In \Cref{sec:exp-exp}, we outline key ideas in our explainable and efficient exploration. In \Cref{sec:k-armed}, we apply them to $K$-armed bandits. In \Cref{sec:lin-bandit}, we address the more general setting of linear bandits with a changing action set. Finally, we show numerical experiments in \Cref{sec:experiments} and conclude in \Cref{sec:conc}.

\section{Problem Formulation}
\label{sec:model}
We start by introducing our general notation. For any $p\geq 0,$ we define a $p$-norm $\|\cdot\|_p$. 
We denote by $\mathbf{1}_k$ the $k$-dimensional vector with all entries equal to $1$. 
We use $\I{\cdot}$ to denote the indicator function. For any positive semi-definite matrix $M$, we let $\|x\|_{M} \coloneqq \sqrt{x^\top M x}$. A $d \times d$ identity matrix is denoted by $\bI_d$. We also define $[m] \coloneqq \{1,2,\ldots,m\}$. 

The bandit setting is defined as follows. The learning horizon has $n$ rounds and we index them by $t \in [n]$. The action set is denoted by $\cA \subseteq \R^d$. It can change with round $t$ and we denote by $\cA_t \subseteq \cA$ the action set in round $t$. A random variable $X$ is $\sigma^2$-sub-Gaussian if $\E[X] = 0$ and satisfies $\E[\exp(sX)]\leq\exp(s^2\sigma^2/2)$ for all $s\in\realset$. The learner interacts with the bandit sequentially. At round $t$, it takes an action $A_t \in \cA_t$ and obtains a reward $X_t=\langle A_t, \btheta_*\rangle + \eta_t,$ where $\eta_t$ is conditionally $1$-sub-Gaussian. The model parameter $\btheta_*\in\R^d$ is unknown to the learner but the features of the actions in $\A_t$ are known. We assume that $\|\btheta_*\|_2\leq S$ and $\|a\|_2 \leq L$ for any $a\in\cA$. The cardinality of each $\cA_t$ is at most $K,$ $|\cA_t|\leq K$.

The goal of the learner is to minimize pseudo-regret
\begin{align*}
R_n &=\E\left[\left(\sum_{t=1}^n\left\langle A_t^*, \btheta_*\right\rangle\right)-\left(\sum_{t=1}^n\left\langle A_t, \btheta_*\right\rangle\right)\right]\,,
\end{align*} 
where $A^*_t = \argmax_{a\in\A_t}\langle a, \btheta_* \rangle$ is the optimal action at round $t$.

\section{Interpretable Exploration via Constrained Optimal Design}
\label{sec:exp-exp}
We introduce \code, a bandit algorithm based on a \textbf{C}onstrained \textbf{O}ptimal \textbf{DE}sign, that is both efficient and interpretable. The key idea in \code is to take the action that maximally reduces the uncertainty in the unknown model parameter $\btheta_*$ from plausible actions. This is formulated as an optimization problem, a variant of the $D$-optimal design subject to constraint. The constraint excludes high-probability sub-optimal actions and defines the plausible action set. In round $t$, \code takes a \textit{deterministic action} that is computed by solving
\begin{align}
  A_t = &\argmax_{a\in\tA_t} \log\det(V_t + a a^\top) \label{eq:d_optimal} \\
  \text { s.t. } \tA_t
  &= \{a \in \A_t:
  \max_{\btheta \in \mathcal{C}_t} \langle a, \btheta\rangle
  \geq \max_{a' \in \A_t}
  \min_{\btheta \in \mathcal{C}_t} \langle a', \btheta\rangle\}\,
  \nonumber
\end{align}

where $V_t\coloneqq\sum_{s=1}^{t-1}A_sA^{\top}_{s}$ is the \emph{design matrix} up to round $t$ that allows computation of high-probability confidence intervals in linear and generalized linear models, and $\C_t$ represents a high-probability confidence set inside which the unknown model parameter $\btheta_*$ resides. We make the following remarks about \code:
\par
    \textbf{(1)} \textbf{Plausible Action Set $\tA_t$:} To handle the possibly changing action set, we introduce a set of per-round constraints that excludes high-probability sub-optimal actions. This is achieved by first building the confidence set $\C_t$ at the beginning of each round $t$. Then we compare the \emph{upper confidence bound (UCB)} of each action $a$, given by $\max_{\btheta \in \mathcal{C}_t}\langle a, \btheta\rangle$, against the \emph{lower confidence bound (LCB)} of any other action $a'$, given by $\min_{\btheta \in \mathcal{C}_t}\langle a', \btheta\rangle$. The constraint ensures that actions with lower UCBs than any LCB are excluded. That is, if $\max_{\btheta \in \mathcal{C}_t}\langle a, \btheta\rangle < \min_{\btheta \in \mathcal{C}_t}\langle a', \btheta\rangle$ for any $a'$, the constraint can be only satisfied by not taking action $a$.

    This is in a sharp contrast to classic elimination algorithms, which eliminate sub-optimal actions from a fixed action set. Our design immediately enables \code to deal with changing action sets. 
    \par
    \textbf{(2)} \textbf{Interpretability:} \code elevates interpretability by design. Specifically, it excludes high-probability sub-optimal actions through the constraint. Then it takes the action, from the remaining plausible actions, that maximally reduces uncertainty in $\btheta_*$, \ie, the one with highest uncertainty in its estimated mean reward. To see this, note that
    \begin{align}
        \argmax_{a\in\tA_t} &\log\det(V_t + a a^\top)\nonumber\\
        & \overset{(a)}{=} \argmax_{a\in\tA_t} \log\det(\bI_d + V_t^{-1/2} a a^\top V_t^{-1/2}) \nonumber\\
        & \overset{(b)}{=} \argmax_{a\in\tA_t} \log(1 + \|a\|^2_{V_t^{-1}})\,.
        \label{eq:equiv}
    \end{align}
    Here equality (a) follows from $V_t$ being constant at round $t$. Equality (b) follows from the fact that for any vector $b$, $\bI_d + b b^\top$ has $d - 1$ eigenvalues $1$ and one eigenvalue $1 + \|b\|_2^2$. Therefore, its determinant is $1 + \|b\|_2^2$ and its logarithm is $\log(1 + \|b\|_2^2)$. By definition, all near-optimal actions are in $\tA_t$ with a high probability, and reducing the uncertainty in their reward estimates increases interpretability. 

\subsection{Explainability Metric}
\label{sec:explainability metric}

To measure an algorithm's interpretability, we introduce the metric of \emph{cumulative model uncertainty error},
\begin{align}
  Q_n
  = \sum_{t = 1}^n \max_{a \in \tA_t} \, (a^\top (\wtheta_t - \btheta_*))^2\,,
  \label{eq:explainability metric}
\end{align}
where $\btheta_*$ is the true unknown model parameter and $\wtheta_t$ is its maximum likelihood estimate from all interactions up to round $t$. We emphasize that, when evaluating an algorithm, the set $\tA_t$ is computed w.r.t. this algorithm's own historical data.

When the value of $Q_n$ is small, it must be true that most $\max_{a \in \tA_t} (a^\top (\wtheta_t - \btheta_*))^2$ are small, and therefore the mean reward estimates of the near-optimal actions in most rounds are close to the corresponding actual mean rewards. As discussed before, this facilitates interpretability, since a human policy designer could act more informatively on a model where the mean rewards of near-optimal actions are estimated more precisely. This could help to dramatically reduce regret for similar MAB problems in the future \citep{garivier2018explore}.

\subsection{Related Optimal Design Works}

Now we discuss some related works that connect our optimization problem in \eqref{eq:d_optimal} to optimal designs. There is a growing body of literature on optimal designs for bandits \citep{pukelsheim2006optimal, fedorov2010optimal, lattimore2020bandit}. The output of optimal designs is a distribution over actions. When the learner takes $n$ actions according to this distribution, it minimizes some notion of uncertainty in the estimate of $\btheta_*$ within a fixed budget of $n$ samples. We note that the optimization in \eqref{eq:d_optimal} maximizes the log determinant of the design matrix, which is similar to the $D$-optimal design. By Kiefer-Wolfowitz theorem \citep{kiefer1960equivalence}, the $D$- and $G$-optimal designs have identical solutions in some problem classes.

Several bandits papers used $G$-optimal designs for regret minimization \citep{soare14bestarm, fiez2019sequential, katz2020empirical, wagenmaker2021experimental, jamieson2022interactive,ZhuK22a}. However, they used them in phases of increasing length and eliminated sub-optimal actions at the end of each phase. The next phase communicates with the earlier one only through non-eliminated actions and neglects all earlier observations. Therefore, these algorithms are conservative and need a fixed action set $\A$. In contrast, \code does not have phases and can be applied to changing action sets $\A_t$. \code also does not compute a sampling distribution over actions at each round $t$, because it takes a plausible deterministic action that maximally reduces the uncertainty in the estimate of $\btheta_*$. We outperform the phased elimination algorithms empirically (\Cref{sec:experiments}). 

Finally, optimal designs have been used in pure exploration and related settings \citep{soare14bestarm, fiez2019sequential, katz2020empirical, fontaine2021online,mason2021nearly,mukherjee2022chernoff,ZhuK22}. We focus on cumulative regret minimization in this work.

\section{$K$-Armed Bandits: A Warm-Up}
\label{sec:k-armed}
To gain preliminary insights into \code, we first study it in classic $K$-armed bandits. We start by introducing $K$-armed bandits and then discuss how \eqref{eq:d_optimal} can be implemented. We represent the $K$-armed bandits as having an action set $\cA = \{\be_1, \dots, \be_K\}\subseteq\R^K$, where $\be_a$ is the $a$-th vector in the standard Euclidean basis representing action $a$. Note that now we use $a$ to denote the action index in $[K]$. We denote by $\mu(a)$ the mean reward of action $a$. The optimal action is $a^* = \argmax_{a \in \cA} \mu(a)$ and the suboptimality gap of action $a$ is $\Delta(a)\coloneqq \mu(a^*) - \mu(a)$.

\code can be implemented as follows. Let
\begin{align*}
  T_t(a) \coloneqq \sum_{s = 1}^{t - 1} \I{A_s = a}   
\end{align*}
be the number of times that action $a$ is taken up to round $t$. Let the mean reward estimate of action $a$ in round $t$ be
\begin{align*}
  \wmu_t(a) \coloneqq T_t^{-1}(a) \sum_{s = 1}^{t - 1} \I{A_s = a} Y_s\,,
\end{align*}
where $Y_s$ is the observed reward at round $s$. Note that the vectors in $\cA$ are orthonormal. Therefore, the constrained optimization in \eqref{eq:d_optimal} reduces to
\begin{align*}
    A_t &\overset{(a)}{=} \argmax_{a\in \tA_t}\log\det(V_t + \be_a \be_a^\top) \\
    \tA_t
  & = \{a \in [K]:
  \max_{\btheta \in \mathcal{C}_t} \langle \be_a, \btheta\rangle
  \geq \max_{a' \in [K]}
  \min_{\btheta \in \mathcal{C}_t} \langle \be_{a'}, \btheta\rangle\}\,
\end{align*}
where the equality (a) holds from \eqref{eq:equiv} and that each action is a vector $\be_a$. It follows then that
\begin{align*}
    A_t &\overset{(a)}{=}\argmax_{a\in\tA_t} \log(1 + \|\be_a\|^2_{V_t^{-1}}) 
    \overset{(b)}{=}  \argmax_{a\in\tA_t} \|\be_a\|^2_{V_t^{-1}}\\
    &\overset{(c)}{=} \argmax_{a\in\tA_t} \dfrac{1}{T_t(a)}\,.
\end{align*} 
Equality (a) holds from \eqref{eq:equiv} and that each action is a vector $\be_a$. Equality (b) follows from the monotonicity of the logarithm. Equality (c) holds because $V_t^{-1}$ is a diagonal matrix whose $a$-th diagonal entry is $T_t^{-1}(a)$.



Now recall that $\mathcal{C}_t$ in the constraint of \eqref{eq:d_optimal} is a confidence set that contains the unknown model parameter with a high probability. 
In our $K$-armed bandits, we have
\begin{align*}
  \max_{\btheta \in \mathcal{C}_t} \langle \be_a, \btheta\rangle
  & = \wmu_t(a) + c_t(a)\,, \\
  \min_{\btheta \in \mathcal{C}_t} \langle \be_{a'}, \btheta\rangle
  & = \wmu_t(a')- c_t(a)\,,
\end{align*} 
where $c_t(a)\coloneqq \sqrt{\frac{2\log(1/\delta)}{T_t(a)}}$  and $\delta \in (0, 1)$ is the failure probability of the bounds. Let $U_t(a) \coloneqq \wmu_t(a) + c_t(a)$ and $L_t(a) \coloneqq \wmu_t(a)-c_t(a)$. With the above, the constrained optimization problem in \eqref{eq:d_optimal} becomes
\begin{align}
    & A_t =  \argmax_{a\in\tA_t} \ \dfrac{1}{T_t(a)}\label{eq:opt-$A$-actioned}\\
    &\mathrm{s.t.} \
    \tA_t = \{a\in [K]: U_t(a) \geq \max_{a'\in[K]} L_t(a')\} \,.\nonumber
\end{align} 
The constraint in \eqref{eq:d_optimal} reduces to comparing the upper and lower confidence bounds of actions $a$ and $a'$, denoted by $U_t(a)$ and $L_t(a')$, respectively \citep{auer2002finite,auer2010ucb}. If $U_t(a)$ is lower than $L_t(a')$ for any action $a'$, then the action $a$ does not belong to the plausible set of actions and cannot be taken. Finally, to avoid division by zero in the objective, each action is initially taken once. 

To summarize, \code solves the optimization in \eqref{eq:opt-$A$-actioned} in every round $t$ and takes a deterministic action. Then it updates the parameters of the taken action $A_t$, its mean reward estimate $\wmu_t(A_t)$, and the number of times that the action is taken $T_t(A_t)$. We analyze the regret of \code next.

\subsection{Regret Analysis}
In this section, we bound the regret of \code in $K$-armed bandits. We state the bound first.

\begin{customtheorem}{1}
\label{thm:k-action}
The regret of \code in $K$-armed bandits is $R_n = O\left(\dfrac{K\log n}{\Delta_{\min}}\right)$, where $\Delta_{\min} \coloneqq \min_{a \in \cA} \Delta(a)$.
\end{customtheorem}

The regret bound in \Cref{thm:k-action} scales linearly with the number of actions $K$ and depends on the minimum gap $\Delta_{\min}$. This is natural in $K$-armed bandits as actions do not share any information. The factor of $\log n$ is standard in gap-dependent regret bounds in $K$-armed bandits \citep{auer2002finite, lattimore2020bandit}. In fact, from Theorem 16.4 of \citet{lattimore2020bandit}, our regret bound matches the gap-dependent lower bound in $K$-armed bandits. This bound also matches that of \linucb and \lints.

\textbf{Gap-independent Bound:} Note that the result of \Cref{thm:k-action} is an instance-dependent bound. We can get an instance-free minimax bound by setting $\Delta_{\min} = \sqrt{K \log n/n}$. It follows then that
\begin{align*}
    R_n = O\left(\sum_{a\in[K]}\Delta(a) +  \dfrac{K\log n}{\Delta_{\min}} \right) =   O(\sqrt{Kn \log n})
\end{align*}
Therefore the gap-independent bound yields a regret of $O(\sqrt{Kn \log n})$.



\textbf{Proof Sketch of \Cref{thm:k-action}:}
We start by considering the good event $\xi_n$ such that the unknown reward parameters fall into the confidence intervals holds for all actions $a\in\A$.  Let $A_t$ be the action selected at round $t$, the $c_t(a) \coloneqq \sqrt{2\log(1/\delta)/T_t(a)}$ be the width of the confidence interval of action $a$, and $a^* \coloneqq \argmax_{a\in\A}\mu(a)$ be the optimal action. The good concentration event is shown in the following lemma.
\begin{lemma}\textbf{(Concentration)}
\label{lemma:k-action-conc} 
Let $\xi_n$ be a good event, that the confidence intervals hold jointly over all rounds and actions,
\begin{align*}
\xi_n \coloneqq \bigcap_{a\in[K]}\bigcap_{t=1}^n\left\{\left|\wmu_{t}(a) - \mu(a)\right| \leq c_{t}(a)\right\}.
\end{align*} 
Then $\Pb(\xi_n) \geq 1-2Kn^2\delta$ for any $\delta\in(0,1)$. 
\end{lemma}
The proof of this lemma is given in \Cref{app:k-armed}. 
From \Cref{lemma:k-action-conc} we know $\xi_n$ holds with probability greater than $(1-2Kn^2\delta)$. Then, we show that, under $\xi_n$, the per-round regret incurred by choosing action according to \eqref{eq:opt-$A$-actioned} and \code is upper bounded by $\mu_{}(a^*) - \mu_{}(A_t) \leq 2c_t(A_t) + 2c_t(a^*)$. This is shown in \Cref{lemma:k-action-per-step}.
\begin{lemma}\textbf{(Per-Round Regret)}
\label{lemma:k-action-per-step}
Suppose event $\xi_n$ occurs and $A_t$ be the action taken by \code in round $t$. Then, $\mu(a^*) - \mu(A_t) \leq 2c_t(A_t) + 2c_t(a^*)$ holds.
\end{lemma}
\begin{proof}
The regret of round $t$ is
\begin{align}\label{eq:k-action-per-round-regret}
\nonumber\mu(a^*)- \mu(A_t) &= (\mu(a^*) - L_t(a^*))\\
&+ (L_t(a^*) - U_t(A_t)) + (U_t(A_t) - \mu(A_t))\nonumber\\
&\leq(\mu(a^*) - L_t(a^*))+(U_t(A_t) - \mu(A_t)),
\end{align} 
where the inequality follows because according to the constraints in \eqref{eq:opt-$A$-actioned}: if $A_t$ is chosen, its UCB $U_t(A_t)$ should be no less than any actions' LCB, and in particular, the optimal action's (\ie, $L_t(a^*)$).

Under $\xi_n$, we have from \cref{lemma:k-action-conc} that
\begin{align*}
\mu(a^*) - L_t(a^*) &= \mu(a^*) - (\wmu_{t}(a^*) - c_t(a^*))\\
&= (\mu(a^*) - \wmu_{t}(a^*) + c_t(a^*)\leq 2c_t(a^*)
\end{align*}
and 
\begin{align*}
U_t(A_t) - \mu(A_t) &= (\wmu_{t}(A_t) + c_t(A_t)) - \mu(A_t)\\
&= (\wmu_{t}(A_t) - \mu(A_t)) +c_t(A_t) \leq 2c_t(A_t).
\end{align*}
Plugging the above two into \eqref{eq:k-action-per-round-regret} finishes the proof.
\end{proof}

There are two key insights in \Cref{lemma:k-action-per-step}. First, to arrive at \eqref{eq:k-action-per-round-regret}, we use the constraints in \eqref{eq:opt-$A$-actioned}, which ensure that the UCB of $A_t$ is at least as large as the LCB of $a^*$. We also observe that the per-round regret depends not only on the taken action $A_t$, but also on the optimal action $a^*$. 


\begin{lemma}\textbf{(Optimal Action Confidence Interval)}
\label{lemma:mu-star}
Let event $\xi_n$ occur and $A_t$ be the action taken by \code in round $t$. Then $\mu(a^*) - \mu(A_t) \leq 4 c_t(A_t)$ holds.
\end{lemma}
\begin{proof}
Now observe from the solution of the constrained optimization in \eqref{eq:opt-$A$-actioned} that it takes those actions that are least taken. This implies that these actions have a higher confidence width than the actions that are taken more. 
Since the action $A_t$ taken at round $t$ has the highest confidence interval width, we have that $c_t(A_t) \geq c_t(a^*)$. Following \cref{lemma:k-action-per-step}, we have $\mu_(a^*) - \mu_(A_t) \leq 4 c_t(A_t)$. The claim of the lemma follows.
\end{proof}
From \Cref{lemma:mu-star} observe that the solution in \eqref{eq:opt-$A$-actioned} ensures that $c_t(A_t) \geq c_t(a^*)$ which leads to the per step regret bound of $4 c_t(A_t)$. 
We finish the proof of \Cref{thm:k-action} by calling \Cref{lemma:k-action-conc}, \ref{lemma:k-action-per-step}, and \ref{lemma:mu-star} and 
showing that the optimal action $a^*$ eliminates sub-optimal actions and bounding the number of samples of each sub-optimal action $a$ till horizon $n$. The full proof is given in \Cref{app:k-armed}. \hfill $\blacksquare$

\subsection{Explainability Metric}
\label{sec:explainability mab}

A $K$-armed bandit can be also used to show that \code minimizes \eqref{eq:explainability metric}. Specifically, based on the formulation in \eqref{eq:opt-$A$-actioned}, \code takes the least taken action among plausible actions $\tA_t$. Since the cardinality of $\tA_t$ is at most $K$, each plausible action must be taken at least $\lfloor (t - 1) / K\rfloor$ times up to round $t$. Taking the form of high-probability confidence intervals $c_t(a)$ into account, we have that
\begin{align*}
  Q_n
  \leq K +
  \sum_{t = K + 1}^n \frac{4 \log(1 / \delta)}{\lfloor (t - 1) / K\rfloor}
  = O(K \log(1 / \delta) \log n)
\end{align*}
holds with probability at least $K n \delta$.

This setting also shows why classic algorithms, like Thompson sampling, do not minimize $Q_n$. Roughly speaking, the plausible actions in TS are all actions because any action can be taken in any round, albeit with a low probability. The unlikely taken actions are highly uncertain by definition and essentially drive $\max_{a \in \cA} \, (a^\top (\wtheta_t - \btheta_*))^2$ in \eqref{eq:explainability metric}.

\section{Linear Bandits}
\label{sec:lin-bandit}
Next, we study \code in linear bandits. Recall that actions in linear bandits are $d$-dimensional vectors $a \in \R^d$ that share the model parameter $\btheta_*$. In round $t$, \code first estimates $\btheta_*$ using regularized least squares from all observations up to round $t$ as
\begin{align}
\wtheta_t\coloneqq\argmin_{\btheta \in \R^d}\left(\sum_{s=1}^{t-1}\left(Y_s-\left\langle\btheta, A_s\right\rangle\right)^2+\lambda\|\btheta\|_2^2\right), \label{eq:theta-hat}
\end{align} 
where $A_s$ is the taken action at round $s$, $Y_s$ denotes its reward, and $\lambda \geq 0$ is a regularizer. Then it uses $\wtheta_t$ to construct a high-probability confidence ellipsoid
\begin{align}
  \C_t
  \coloneqq \{\btheta \in \realset^d: \|\wtheta_t - \btheta\|_{V_t}
  \leq \sqrt{\beta_t(\delta)}\}\,,
  \label{eq:ellipsoid}
\end{align}
where $\delta \in (0, 1)$ is its failure probability and $\sqrt{\beta_t(\delta)} = \sqrt{d \log((1 + t L^2 / \lambda) \delta)} + \lambda^{1 / 2} S$. Finally, it solves the optimization in \eqref{eq:d_optimal} to select $A_t$ and observes its reward $Y_t$. The pseudo-code is presented in \Cref{alg:linear-bandit}. 

\begin{algorithm}[!tbh]
\caption{\code\ for linear bandits.}
\label{alg:linear-bandit}
\begin{algorithmic}[1]
\STATE Input: Finite action set $\cA \subseteq \mathbb{R}^d$, regularizer $\lambda \geq 0$
\STATE Set $V_0 = \lambda \bI_d$
\FOR{$t = 1, \dots, n$}
\STATE Estimate $\wtheta_t$ using \eqref{eq:theta-hat} and $C_t$ using \eqref{eq:ellipsoid}
\STATE Solve the optimization in \eqref{eq:d_optimal}
\STATE Select action $A_t$ and observe reward $Y_t$
\ENDFOR
\end{algorithmic}
\end{algorithm}

\subsection{Regret Analysis}

In this section, we bound the regret of \code in linear bandits. We state the bound first.

\begin{customtheorem}{2}
\label{thm:lin-bandit}
The regret of \code\ in linear bandits is given by
$R_n=\widetilde{O}(d\sqrt{n})$, where $\widetilde{O}$ hides logarithmic factors.
\end{customtheorem}

This rate is standard in linear bandits \citep{abbasi2011improved,lattimore2020bandit}, and comparable to those of \linucb and \lints with infinitely many actions. From Theorem 24.2 of \citet{lattimore2020bandit}, our regret bound matches the gap-free lower bound of $\Omega(d\sqrt{n})$ in this setting. Finally, we note that one $\sqrt{d}$ in our bound could be exchanged for $\sqrt{\log K}$ if the confidence ellipsoid in \eqref{eq:ellipsoid} was designed for finitely many actions.

\textbf{Proof Sketch of \Cref{thm:lin-bandit}:}
The regret bound proof follows a similar rationale as \Cref{thm:k-action} with a few novel techniques. We start off by considering the good event $\xi_n$ such that $\btheta_*\in \C_t$ for all $t\in[n]$ as follows:
    $\xi_n \coloneqq \bigcap_{t=1}^n\{\btheta_*\in \C_t\}$.
%
In the following lemma, we show that the good event $\xi_n$ holds with high probability. 
\begin{lemma}\textbf{(Confidence Ellipsoid, Theorem 2 of \citet{abbasi2011improved})}
\label{lemma:supp-2}
    Let $V_0=\lambda \mathbf{I}_d, \lambda>0$. Assume that $\left\|\btheta_*\right\|_2 \leq S$ and $\left\|A_t\right\|_2 \leq L$. Then, for any $\delta>0$, with probability at least $1-\delta$, for all $t \geq 0, \btheta_*$ lies in the confidence set $\C_t$ defined in \eqref{eq:ellipsoid}.
\end{lemma}
The proof of this lemma is in \citet{abbasi2011improved}. It follows from \Cref{lemma:supp-2} that $\Pb(\xi_n)\geq 1 - \delta$. 
Then we provide a key lemma that upper bounds the per-step regret.

\begin{lemma}\textbf{(Per-Round Regret)}
\label{lemma:per-step-regret}
    Under the good event $\xi_n$ we can show that the expected regret of round $t$ is bounded from above as
    \begin{align*}
        &\langle A_t^*,\btheta_*\rangle- \langle A_t,\btheta_*\rangle \\
        \leq &\left(\langle A_t^*,\btheta_*\rangle-\min_{\btheta\in\cC_t}\langle A_t^*,\btheta\rangle\right)+\left(\max_{\btheta\in\cC_t}\langle A_t,\btheta\rangle-\langle A_t,\btheta_*\rangle\right).
    \end{align*}
\end{lemma}
The proof of this lemma is similar to that of \cref{lemma:k-action-per-step}, and is deferred to \cref{sec:lemma:per-step-regret}. 
Following conventional linear bandits analysis, the second term on the R.H.S. of \cref{lemma:per-step-regret} can be bounded as 
\begin{align}
\nonumber\max_{\btheta\in\cC_t}\langle A_t,\btheta\rangle-\langle A_t,\btheta_*\rangle\leq&\|A_t\|_{V^{-1}_t}\max_{\btheta\in\cC_t}\|\btheta-\btheta_*\|_{V_t}\\
\label{eq:code_regret1}\leq&\|A_t\|_{V_t^{-1}}\sqrt{\beta_t(\delta)},\end{align}
where $\beta_t(\delta)$ is the width of $\cC_t$, the confidence region of $\btheta$. 
For the first term of the R.H.S. in \cref{lemma:per-step-regret}, we can similarly have 
\begin{align}
\nonumber\langle A^*_t,\btheta\rangle-\min_{\btheta\in\cC_t}\langle A^*_t,\btheta_*\rangle\leq&\|A_t^*\|_{V^{-1}_t}\max_{\btheta\in\cC_t}\|\btheta-\btheta_*\|_{V_t}\\
\nonumber\leq&\|A^*_t\|_{V_t^{-1}}\sqrt{\beta_t(\delta)}\\
\leq&\|A_t\|_{V_t^{-1}}\sqrt{\beta_t(\delta)}
\label{eq:code_regret2},\end{align}
where the last step holds by definition of \code.

Combining the above, we have that the per round regret of \code~ under the good event $\xi_n$ is at most $2\|A_t\|_{V_t^{-1}}\sqrt{\beta_t(\delta)}$. From here, one can apply the log determinant inequality in \Cref{lemma:supp-1} to give us the desired regret bound. The full proof is given in \Cref{app:lin-bandit}. 

We remark that \code builds the per-round constraint set $\tA_t$ at every round $t$ that excludes sub-optimal actions with high probability. In contrast, Theroem 22.1 of \citet{lattimore2020bandit} (for phased based elimination Algorithm 12) keeps eliminating high probability sub-optimal actions from a fixed action set. Hence, their algorithm (and the proof technique) cannot handle changing action set.


\section{Experiments}
\label{sec:experiments}
We conduct five experiments. In the first experiment, we evaluate \code on synthetic linear bandits and compare it to multiple baselines. 
In the second experiment we compare \code on synthetic linear bandits with changing arms and compare it to the same set of baselines.
The remaining three experiments are conducted on the MovieLens, White Wine Quality, and Heart Failure datasets. We use these standard public benchmarks to evaluate the performance of \code in real-life datasets.

The cumulative model uncertainty is calculated following \eqref{eq:explainability metric}. 
For all the algorithms, we set $\tA_t$ to the actions as stated in the constraint in \eqref{eq:d_optimal} w.r.t. their historical observations.


\subsection{Linear Bandits}
\label{sec:linear bandit experiment}

We start with $d$-dimensional linear bandits, where $d = 5$, the number of actions is $K = 100$, and the horizon is $n = 10\,000$ rounds. The model parameter is generated as $\btheta_* \sim \mathcal{N}(\mathbf{0}_d, I_d)$. The feature vectors of actions are drawn uniformly at random from $[-1, 1]^d$. The reward noise is $\mathcal{N}(0, \sigma^2)$ and $\sigma = 0.5$. All results are averaged over $200$ random runs.

We compare \code to five baselines. The first two, \linucb \citep{dani08stochastic,abbasi2011improved} and \lints \citep{chapelle11empirical,agrawal2013thompson}, are classic adaptive algorithms for linear bandits. They are not interpretable and show what regret is attainable without demanding interpretability. \elim is an elimination algorithm for linear bandits with quadrupling phase lengths (Section 22 in \citet{lattimore2020bandit}). Like \code, \elim uses an optimal design to explore. Unlike \code, the design is computed only at the beginning of each phase and kept fixed throughout it. The \egreedy policy explores in round $t$ with probability $\varepsilon \sqrt{n / t} / 2$. Therefore, it explores more initially and the number of exploration rounds is roughly $\varepsilon n$. The last baseline is explore-then-commit (\etc) of \citet{langford08epochgreedy}. \etc explores in the first $\varepsilon n$ rounds and then acts greedily with respect to the learned model.

We implement \code with the confidence interval in \eqref{eq:ellipsoid}, where $L = 1$, $S = 0$, $\lambda = 10^4$, and $\delta = 0.05$. This setting gives us a comparable performance to \linucb. We set $\delta = 0.05$ in \elim, and $\varepsilon = 0.05$ in both the \egreedy policy and \etc. 

\begin{figure}[t]
  \centering
  \includegraphics[width=3.3in]{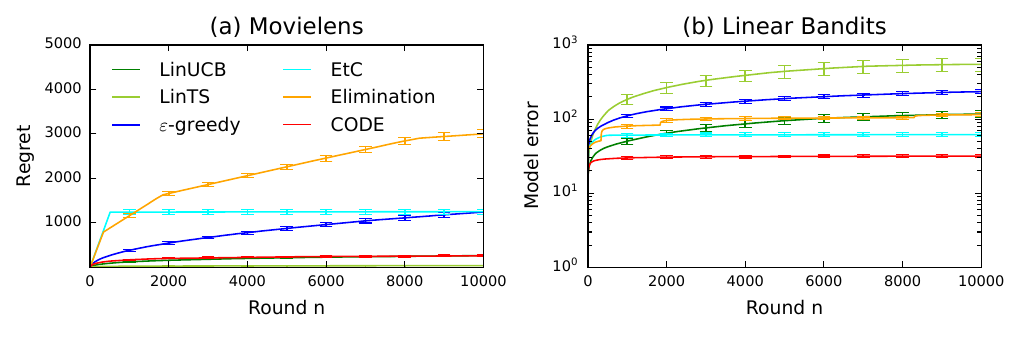}
  \vspace{-0.25in}
  \caption{(a) Regret of \code and five baselines in linear bandits, reported as a function of horizon $n$. (b) Interpretability of \code and the baselines.}
  \label{fig:linear bandit}
\end{figure}

In \cref{fig:linear bandit}a, we plot the regret of all methods as a function of the number of rounds $n$. We observe that the regret of \code is comparable to that of \linucb, and significantly lower than that of any interpretable baseline. \lints, which is a strong but uninterpretable baseline, outperforms all methods. In \cref{fig:linear bandit}b, we plot the \emph{uncertainty loss} due to not acting like the $D$-optimal design. We observe that \code has the lowest uncertainty loss and therefore can be viewed as most interpretable. The next two most interpretable methods are \elim and \etc. This experiment clearly demonstrates the benefit of \code over other interpretable methods.

\subsection{Linear Bandits with Changing Arms}
\label{sec:linear bandit experiment changing}

We now implement a $d$-dimensional linear bandits, where $d = 8$, the number of actions is $K = 200$, and the horizon is $n = 10\,000$ rounds. The model parameter is generated as $\btheta_* \sim \mathcal{N}(\mathbf{0}_d, I_d)$. The feature vectors of actions are drawn uniformly at random from $[-1, 1]^d$. Note that now the actions at every round is uniform randomly chosen, which gives rise to the changing action set. The reward noise is $\mathcal{N}(0, \sigma^2)$ and $\sigma = 0.5$. All results are averaged over $200$ random runs.

We again compare \code to the five baselines: \linucb, \lints, \elim,  \egreedy, and \etc. We implement \code with the confidence interval in \eqref{eq:ellipsoid}, where $L = 1$, $S = 0$, $\lambda = 10^4$, and $\delta = 0.05$. This setting gives us a comparable performance to \linucb. We set $\delta = 0.05$ in \elim, and $\varepsilon = 0.05$ in both the \egreedy policy and \etc. 

\begin{figure}[t]
  \centering
  \includegraphics[width=3.3in]{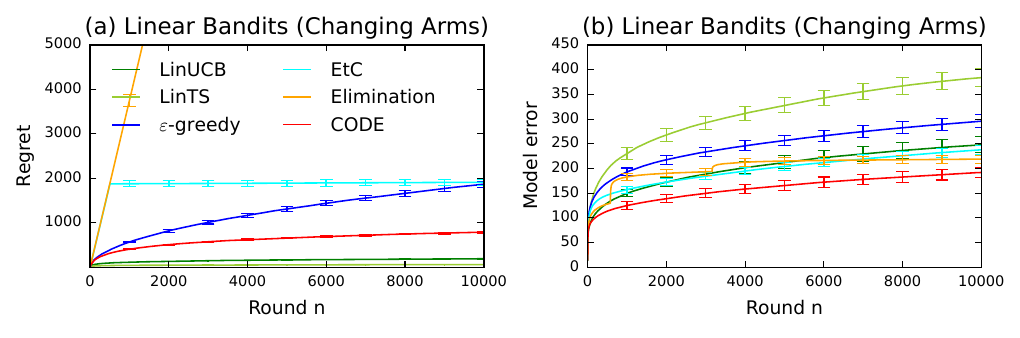}
  \vspace{-0.25in}
  \caption{(a) Regret of \code and five baselines in linear bandits, reported as a function of horizon $n$. (b) Interpretability of \code and the baselines.}
  \label{fig:linear bandit changing}
\end{figure}

In \cref{fig:linear bandit}a, we plot the regret of all methods as a function of the number of rounds $n$. We observe that the regret of \code is comparable to that of \linucb, and significantly lower than that of any interpretable baseline. \lints, which is a strong but uninterpretable baseline, outperforms all methods. In \cref{fig:linear bandit}b, we plot the \emph{uncertainty loss} due to not acting like the $D$-optimal design. We observe that \code has the lowest uncertainty loss and therefore can be viewed as most interpretable. The next two most interpretable methods are \elim and \etc. This experiment clearly demonstrates the benefit of \code over other interpretable methods.

\subsection{MovieLens}
\label{sec:movielens experiment}

We also experiment with the MovieLens 1M dataset \citep{movielens}. This dataset contains one million ratings given by $6\,040$ users to $3\,952$ movies. We apply low-rank factorization to the rating matrix to obtain $5$-dimensional representations: $\btheta_j \in \realset^5$ for user $j \in [6\,040]$ and $a_i \in \realset^5$ for movie $i \in [3\,952]$.  In each run, we randomly select one user $\btheta_j$ and $100$ movies $a_i$, and they represent the unknown model parameter and actions, respectively. The reward noise is $\mathcal{N}(0, \sigma^2)$, where $\sigma = 0.811$ is estimated from data. All results are averaged over $200$ random runs.

\begin{figure}[t]
  \centering
  \includegraphics[width=3.3in]{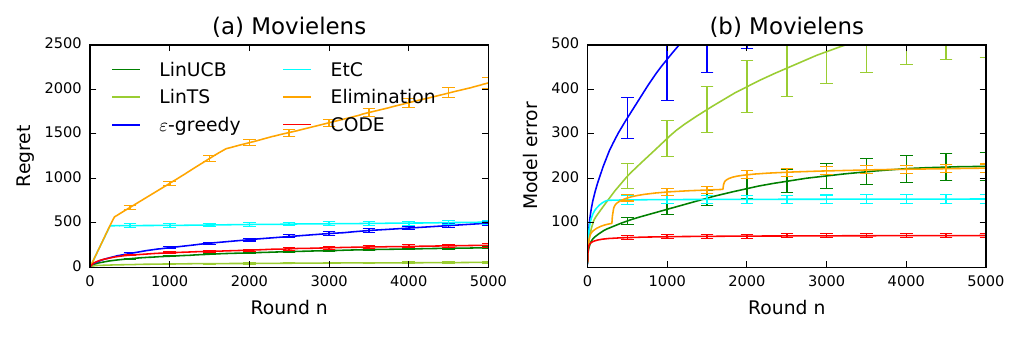}
  \vspace{-0.25in}
  \caption{(a) Regret of \code and five baselines on the MovieLens problem, reported as a function of horizon $n$. (b) Interpretability of \code and the baselines.}
  \label{fig:movielens}
\end{figure}

The performance of all compared methods is reported in \cref{fig:movielens}. We observe the same trends as in \cref{fig:linear bandit}. In \cref{fig:movielens}a, \code has a comparable regret to \linucb and is outperformed only by \lints. In \cref{fig:movielens}b, \code attains the lowest uncertainty loss and thus can be viewed as most interpretable.

\subsection{White Wine Quality}

The White Wine Quality dataset \citep{cortez2009modeling} contains $4\,500$ samples of white wine, each associated with a feature vector $a \in \realset^{11}$. The target variable is the rating of wine in range $1$ to $10$. We first fit a linear model to the dataset and get an estimate of $\btheta_*$. The reward model is linear, $A_t^\top \btheta_* + \eta_t$, where $A_t$ is the action at round $t$ and $\eta_t$ is conditional $1$-sub-Gaussian noise. In each run, we randomly select $100$ wines as actions. Our results are averaged over $50$ runs.

\begin{figure}[t]
  \centering
  \includegraphics[width=3.3in]{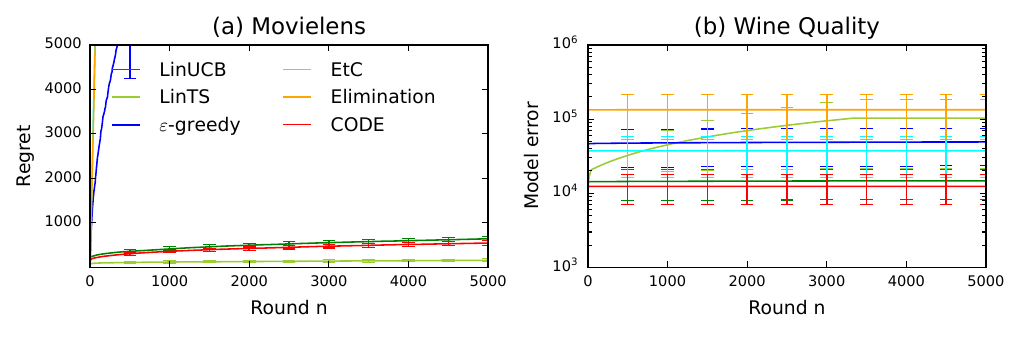}
  \vspace{-0.25in}
  \caption{(a) Regret of \code and five baselines on the Wine Quality dataset, shown as a function of horizon $n$. (b) Interpretability of \code and the baselines.}
  \label{fig:wine}
\end{figure}

The performance of all compared methods is reported in \cref{fig:wine}. We observe again that \lints is the only baseline with a significantly lower regret than \code (\cref{fig:wine}a). On the other hand, \code is most interpretable (\cref{fig:wine}b).

\subsection{Heart Failure}

The Heart Failure dataset \citep{misc_heart_failure_clinical_records_519} contains $300$ samples of heart failure records, each associated with a feature vector $a \in \realset^{12}$. The target variable is if the patient lived or died. We again fit a linear model to the dataset and get an estimate of $\btheta_*$. The reward model is linear, $A_t^\top \btheta_* + \eta_t$, where $A_t$ is the action at round $t$ and $\eta_t$ is conditional $1$-sub-Gaussian noise. All samples are actions. Our results are averaged over $50$ runs.

\begin{figure}[t]
  \centering
  \includegraphics[width=3.3in]{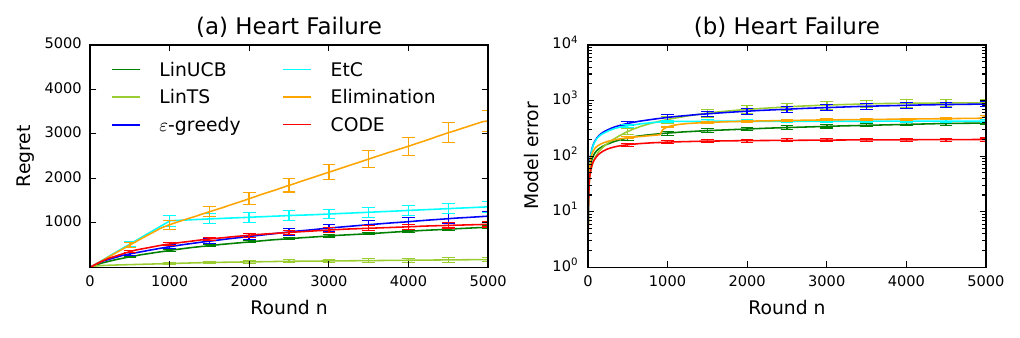}
  \vspace{-0.25in}
  \caption{(a) Regret of \code and five baselines on the Heart Failure dataset, shown as a function of horizon $n$. (b) Interpretability of \code and the baselines.}
  \label{fig:heart}
  \vspace{-0.2in}
\end{figure}

The performance of all compared methods is reported in \cref{fig:heart}. We observe that \lints has the lowest regret, followed by all other methods but \elim (\cref{fig:heart}a). On the other hand, \code is most interpretable by a large margin (\cref{fig:heart}b).

\section{Conclusions and Future Work}
\label{sec:conc}
We propose algorithm \code for linear bandits that is computationally efficient, statistically efficient, and interpretable. The key idea in \code is to take the action that maximally reduces the uncertainty in the unknown model parameter, from potentially optimal actions. Therefore, \code is comparably simple to other interpretable algorithms, such as \etc and \egreedy. \code can also be viewed as removing phases in phased elimination algorithms. It addresses many issues that plagued these designs \citep{soare14bestarm,fiez2019sequential} and affected their practical performance. We analyze \code in $K$-armed and linear bandits, and prove tight regret bounds. Finally, we empirically evaluate \code and compare it to state-of-the-art baselines in linear bandits. \code outperforms all interpretable baselines, while matching the performance of popular but uninterpretable designs like \linucb. 
In the future work, we want to extend this work to a more general class of reinforcement learning (RL) problems, such as regret minimization in contextual bandits and linear MDPs.



\bibliographystyle{plainnat}
\bibliography{biblio}

\newpage
\onecolumn
\appendix
\section{Appendix}
\subsection{$K$-armed Bandit Proof}
\label{app:k-armed}
\begin{customlemma}{1}\textbf{(Restatement of Concentration lemma)}
Define the confidence interval for each action $a\in[K]$ at round $t$ as $c_t(a) = \sqrt{\frac{2\log(1/\delta)}{T_t(a)}}$. 
Let $\xi_n$ be the good event that confidence intervals hold for all $n$ such that 
\begin{align*}
\xi_n \coloneqq \bigcap_{a\in\A}\bigcap_{t=1}^n\left\{\left|\wmu_{t}(a) - \mu(a)\right| \leq c_{t}(a)\right\}.
\end{align*} 
Then the $\Pb(\xi_n) \geq 1-2n^2\delta$ for any $\delta\in(0,1)$. 
\end{customlemma}

\begin{proof}
We can show that for the action $a$ the event $\xi_{n}(a)$ not happening is bounded by the probability of the following three bad events
\begin{align}
    \bigcup_{t=1}^n\{\xi^c_t\} &\subseteq  \bigcup_{t=1}^n\underbrace{\{\wmu_{t}(a) \geq \mu(a) + c_{t}(a)\}}_{\textbf{action $a$ is over-estimated}} \bigcup \underbrace{\{\wmu_{t}(a^*) \leq \mu(a^*) - c_{t}(a^*)\}}_{\textbf{action $a^*$ is under-estimated}} \nonumber\\
    &\qquad \bigcup \underbrace{\{ \mu_{}(a^*) - \mu_{}(a) \leq 2 c_{t}(a)\}}_{\textbf{Critical number of samples}} \label{eq:bad-event-decomposition}
\end{align} 
where, $t=1,2,\ldots,n$. 
Then using Hoeffding inequality for bounded rewards and union bounding over all $t=1,2,\ldots,n$ we can show that
\begin{align}
    \Pb\left(\bigcup_{a\in\A}\bigcup_{t=1}^n\{\wmu_{t}(a) \geq \mu(a) + c_{t}(a)\}\right) &\leq \sum_{a\in\A}\sum_{t=1}^n\sum_{T_t(a)=1}^t \Pb\left(\{\wmu_{t}(a) \geq \mu(a) + c_{t}(a)\}\right)\nonumber\\
    &\overset{(a)}{\leq} \sum_{a\in\A}\sum_{t=1}^n\sum_{T_t(a)=1}^t\exp\left(-\frac{c^2_t(a) T_t(a)}{2}\right) = Kn^2\delta
    \label{eq:-bad-elim-1}
\end{align}
where, $(a)$ follows from Hoeffding's inequality and using the $1$-sub-Gaussian assumption.
Similarly for the other event also we can show that
\begin{align}
    \Pb\left(\sum_{a\in\A}\bigcup_{t=1}^n\{\wmu_{t}(a^*) \leq \mu(a^*) - c_{t}(a^*)\}\right) \leq  Kn^2\delta \label{eq:-bad-elim-2}
\end{align}
Finally, for the third bad event, we can show that
\begin{align}
    \mu_{}(a^*) - \mu_{}(a) \leq 2 c_{t}(a) \implies T_{t}(a) \leq   \dfrac{8\log n}{\Delta^2(a)} \label{eq:-bad-elim-3}
\end{align} 
Hence, for $T_{t}(a) \geq  1 +  \frac{8\log n}{\Delta^2(a)}$ the third bad event is controlled. This we call the critical number of samples. Hence, combining \eqref{eq:-bad-elim-1}, \eqref{eq:-bad-elim-2}, and \eqref{eq:-bad-elim-3} we can show that $\Pb(\xi_{n}(a)) \geq 1 - 2K n^2\delta$ for $T_n(a)\geq 1 + \frac{8\log n}{\Delta^2(a)}$.
\end{proof}

\begin{customtheorem}{1}\textbf{(Restatement)}
The regret of the \code\ in $K$-actioned bandit 
is given by $R_n = O\left(\dfrac{K\log n}{\Delta_{\min}}\right)$, where $\Delta_{\min} = \min_a\Delta(a)$.  
\end{customtheorem}

\begin{proof}
\textbf{Step 1 (Definition):} Define the confidence width for an action $a$ at round $t$ as $c_t(a) \coloneqq \sqrt{\frac{2\log (1 / \delta)}{T_t(a)}}$. 
Define the LCB of action $a$ as $L_t(a) \coloneqq \wmu_t(a) - c_t(a)$ and the UCB as $U_t(a) \coloneqq \wmu_t(a) + c_t(a)$ at round $t$. Let the empirical mean for action $a$ till round $t$ be defined as $\wmu_t(a)$.
Let $\ell_t = \argmax_{a\in\A} L_t(a)$ 
be the index of the action with the highest LCB. Define the good event $\xi_n$ such that all confidence interval holds for all actions $a\in\A$ as follows:
\begin{align*}
\xi_n \coloneqq \bigcap_{a\in [K]}\bigcap_{t=1}^n\left\{\left|\wmu_{t}(a) - \mu(a)\right| \leq c_{t}(a)\right\}.
\end{align*} 
From \Cref{lemma:k-action-conc} we know that the probability of the good event is bounded by at least $(1-2Kn\delta)$.

\textbf{Step 2 (Per step concentration):} Under the good event  $\xi_n$ and sampling according to \eqref{eq:opt-$A$-actioned} we can show that the $\mu_{}(a^*) - \mu_{}(A_t) \leq 4 c_t(A_t)$. This follows from \Cref{lemma:k-action-per-step} and \Cref{lemma:mu-star}.

\textbf{Step 3 (Bound the samples of sub-optimal actions $a\in\A\setminus\{a^*\}$):} First recall that under the good event $\xi_n$ an action $a$ can only be taken at round $t$ if 
\begin{align*}
    L_t(a^*) \leq U_t(a) \overset{(a)}{\implies} \mu(a^*) - \mu_{}(a_t) \leq 4 c_t(a_t) &\overset{(b)}{\implies} \Delta(a) \leq \sqrt{\dfrac{2\log(1/\delta)}{T_t(a)}}\\
    & \implies T_t(a) \leq \dfrac{2\log(1/\delta)}{\Delta^2(a)}
\end{align*}
where, $(a)$ follows from step 4, $(b)$ follows from the definition of the gap. Now to get the total number of times the action $a$ is taken, we can sum up $T_t(a)$ from $t=1$ to $\frac{2 \log(1 / \delta) }{\Delta^2(a)}$. This yields

\begin{align*}
    \sum_{t=1}^{\dfrac{2 \log(1 / \delta) }{\Delta^2(a)}}\sqrt{\dfrac{2 \log(1 / \delta)}{t}} \leq \sqrt{2\log(1/\delta)}\int_{t=1}^{\dfrac{2 \log(1 / \delta) }{\Delta^2(a)}}\dfrac{1}{\sqrt{t}} &\overset{(a)}{\leq}\sqrt{2\log(1/\delta)}\left(2\sqrt{\dfrac{2 \log(1 / \delta)}{\Delta^2(a)}}\right)\\
    &\leq \dfrac{8 \log(1/\delta)}{\Delta(a)} 
\end{align*}
where, $(a)$ follows as $\int_1^b \frac{1}{\sqrt{t}} d t=2 \sqrt{b}-2$.

\textbf{Step 4 (Regret Decomposition):} Finally we can decompose the regret till $n$ as follows:
\begin{align*}
    R_n = \sum_{a=1}^K \Delta(a)\E[T_n(a)] &= \sum_{a=1}^K \Delta(a)\E[T_n(a) \indic{\xi_n}] + \sum_{a=1}^K \Delta(a)\E[T_n(a) \indic{\xi^C_n}] \\
    &\leq \sum_{a=1}^K \Delta(a)\E[T_n(a) \indic{\xi_n}] + \sum_{a=1}^K \Delta(a)\E[ \indic{\xi^C_n}] \\
    &\leq  \sum_{a=1}^K \Delta(a)\left(1 +  \frac{8\log n}{\Delta^2(a)}\right) + 2\sum_{a=1}^K\Delta(a)n\delta\\
    &= \sum_{a=1}^K \Delta(a) +  \sum_{a=1}^K \frac{8\log (1/\delta)}{\Delta(a)}\ + 2\sum_{a=1}^K\Delta(a)n\delta 
\end{align*}
The claim of the theorem follows by setting $\delta = \frac{1}{n^3}$.

\end{proof}

\subsection{Linear Bandit Proof}
\label{app:lin-bandit}
\begin{lemma}\textbf{(Restatement of Lemma 19.4 from \citet{lattimore2020bandit} )}
\label{lemma:supp-1}
Let $A_1, \ldots, A_n \in \mathbb{R}^d$ be a sequence of vectors with $\left\|A_t\right\|_2 \leq L<\infty$ for all $t \in[n]$, $V_t=\sum_{s=1}^{t-1}A_sA^{\top}_s+\lambda \mathbf{I}_d$ with $\lambda\geq L^2$, then
\begin{align*}    \sum_{t=1}^n\left\|A_t\right\|_{V_{t}^{-1}}^2 \leq 2 \log \left(\frac{\operatorname{det} V_{n+1}}{\operatorname{det} V_1}\right) \leq 2 d \log \left(\frac{\operatorname{trace} V_1+n L^2}{d \operatorname{det}\left(V_1\right)^{1 / d}}\right)=2d\log\left( d+nL^2/\lambda\right).
\end{align*}
\end{lemma}

\begin{customtheorem}{2}\textbf{(Restatement)}
The regret of \code\ in the linear bandits setting is given by
\begin{align*}
    R_n\leq 4(2+L) \sqrt{(1 + LS) n d \log (\lambda+n L / d)}\left(\lambda^{1 / 2} S+ \sqrt{2 \log (1 / \delta)+d \log (1+n L /(\lambda d))}\right)
\end{align*} 
where, $\left\|a\right\|_2 \leq L$ for any $a\in\A_t$ and $\|\btheta_*\|_2\leq S$.
\end{customtheorem}

\begin{proof}
    \textbf{Step 1 (Definition)}: Following the work of \citet{abbasi2011improved} we define the width of the confidence interval as 
    \begin{align*}
        \beta_t(\delta) \coloneqq R \sqrt{d \log \left(\frac{1+t L^2 / \lambda}{\delta}\right)}+\lambda^{1 / 2} S.
    \end{align*}
    Then define the confidence set (ellipsoid) at round $t$ as follows:
    \begin{align*}
        \C_{t}\coloneqq\left\{\btheta \in \mathbb{R}^d:\left\|\wtheta_t-\btheta\right\|_{V_t} \leq R \sqrt{d \log \left(\frac{1+t L^2 / \lambda}{\delta}\right)}+\lambda^{1 / 2} S\right\}.
    \end{align*}
    Finally, define the good event that $\btheta_*\in \C_t$ for all $t\in[n]$ as follows:
    \begin{align*}
        \xi_n = \bigcap_{t=1}^n\{\btheta_*\in \C_t\}
    \end{align*}
We can show from \Cref{lemma:supp-2} that $\Pb(\xi_n)\geq 1 - \delta$. 
%
%
Finally, for convenience, we state the algorithm once more as follows:
\begin{align}
  A_t =& \argmax_{a\in\A} \  \log(1 + \|a\|^2_{V_t^{-1}})  \label{eq:d_optimal_lin_bandit-1}\\
  &\mathrm{s.t.} \ \left[\underbrace{\langle a',\hat{\btheta}_t\rangle + \sqrt{2\|a'\|^2_{V^{-1}_t}\log(\delta^{-1})}}_{U_t(a')} - \left(\underbrace{\langle a,\hat{\btheta}_t\rangle-\sqrt{2\|a\|^2_{V^{-1}_t}\log(\delta^{-1})}}_{L_t(a)}\right)\right]\geq 0\quad\forall a,a'\in\cA\,.\nonumber
\end{align}
where $U_t(a)$ and $L_t(a)$ denote the UCB and LCB of the action $a$. Let $\ell_t = \argmax_a L_t(a)$ be the index of the action with the highest LCB. We also define the confidence width of action $a$ as $c_t(a) = \sqrt{2\|a\|^2_{V^{-1}_t}\log(\delta^{-1})}$.

\textbf{Step 2 (Per-step Regret):} Using \cref{lemma:per-step-regret} we can show that the per step regret of round $t$ is bounded as follows
\begin{align}
    &\langle A_t^*,\btheta_*\rangle- \langle A_t,\btheta_*\rangle 
        \leq \left(\langle A_t^*,\btheta_*\rangle-\min_{\btheta\in\cC_t}\langle A_t^*,\btheta\rangle\right)+\left(\max_{\btheta\in\cC_t}\langle A_t,\btheta\rangle-\langle A_t,\btheta_*\rangle\right).\label{eq:lin-ci}
\end{align}
%
%

\textbf{Step 3 (First and Second Term of \eqref{eq:lin-ci}):} The second term of \eqref{eq:lin-ci} can be upper-bounded as 
\begin{align}
\nonumber\max_{\btheta\in\cC_t}\langle A_t,\btheta\rangle-\langle A_t,\btheta_*\rangle\leq&\|A_t\|_{V^{-1}_t}\max_{\btheta\in\cC_t}\|\btheta-\btheta_*\|_{V_t}\\
\label{eq:code_regret1-app}\leq&\|A_t\|_{V_t^{-1}}\sqrt{\beta_t(\delta)},\end{align}
where $\beta_t(\delta)$ is the width of $\cC_t$, the confidence region of $\btheta$.

For the first term of \eqref{eq:lin-ci}, we can similarly have 
\begin{align}
\nonumber\langle A^*_t,\btheta\rangle-\min_{\btheta\in\cC_t}\langle A^*_t,\btheta_*\rangle\leq&\|A_t^*\|_{V^{-1}_t}\max_{\btheta\in\cC_t}\|\btheta-\btheta_*\|_{V_t}\\
\nonumber\leq&\|A^*_t\|_{V_t^{-1}}\sqrt{\beta_t(\delta)}\\
\leq&\|A_t\|_{V_t^{-1}}\sqrt{\beta_t(\delta)}
\label{eq:code_regret2-app},\end{align}
where the last step holds by definition of \code.

%

\textbf{Step 4 (Regret Decomposition):}  Combining the above and putting them back to \eqref{eq:lin-ci}, we have the expected regret of a round $t$ is at most upper bounded by
\begin{align}
    r_t \leq 2\|a\|_{V_t^{-1}}\sqrt{\beta_t(\delta) }.
\end{align}
%
Sum of this over $t=1,\ldots,n$ yields the following:
$$
2\sum_{t=1}^n\|A_t\|_{V_t^{-1}}\sqrt{\beta_t(\delta) }.
$$
We can apply the log determinant inequality in \Cref{lemma:supp-1} from here.
Thus, setting $\lambda > 0 $, with probability at least $1-\delta$, for all $n \geq 0$
\begin{align*}
R_n & \leq \sqrt{n \sum_{t=1}^n r_t^2} \leq \sqrt{8 \beta_n(\delta) n \sum_{t=1}^n\|A_t\|^2_{V_t^{-1}})} \\
&\leq  8 \sqrt{\beta_n(\delta) n \log \left(\operatorname{det}\left(V_n\right)\right)}\\
& \leq 8 \sqrt{ n d \log (\lambda+n L / d)}\left(\lambda^{1 / 2} S+R \sqrt{2 \log (1 / \delta)+d \log (1+n L /(\lambda d))}\right)
\end{align*}
where the last two steps follow from \Cref{lemma:supp-1}.
\end{proof}

\section{Proof of \cref{lemma:per-step-regret}}
\label{sec:lemma:per-step-regret}

\begin{proof}
 We can show that the expected regret of round $t$ is 
\begin{align}
    &\nonumber\langle A_t^*,\btheta_*\rangle- \langle A_t,\btheta_*\rangle\\
    =&\nonumber\langle A_t^*,\btheta_*\rangle-\min_{\btheta\in\cC_t}\langle A^*_t,\btheta\rangle+\min_{\btheta\in\cC_t}\langle A_t^*,\btheta\rangle-\left(\max_{\btheta\in\cC_t}\langle A_t,\btheta\rangle-\max_{\btheta\in\cC_t}\langle A_t,\btheta\rangle+\langle A_t,\btheta_*\rangle\right)\\
    =&\nonumber\langle A_t^*,\btheta_*\rangle-\min_{\btheta\in\cC_t}\langle A^*_t,\btheta\rangle + \underbrace{\min_{\btheta\in\cC_t}\langle A_t^*,\btheta\rangle-\max_{\btheta\in\cC_t}\langle A_t,\btheta\rangle}_{\textbf{Constraint}}-\left(-\max_{\btheta\in\cC_t}\langle A_t,\btheta\rangle+\langle A_t,\btheta_*\rangle\right)\\
    \leq&\langle A_t^*,\btheta_*\rangle-\min_{\btheta\in\cC_t}\langle A_t^*,\btheta\rangle-\left(-\max_{\btheta\in\cC_t}\langle A_t,\btheta\rangle+\langle A_t,\btheta_*\rangle\right)\nonumber\\
    =&\langle A_t^*,\btheta_*\rangle-\min_{\btheta\in\cC_t}\langle A_t^*,\btheta\rangle+\left(\max_{\btheta\in\cC_t}\langle A_t,\btheta\rangle-\langle A_t,\btheta_*\rangle\right).\label{eq:lin-ci-main}
\end{align}
The claim of the lemma follows. 
\end{proof}

\newpage
\section{Table of Notations}
\label{table-notations}

\begin{table}[!tbh]
    \centering
    \begin{tabular}{|p{26em}|p{21em}|}
        \hline\textbf{Notations} & \textbf{Definition} \\\hline
        $K$ & Total number of arms \\\hline
        $d$ & Dimension of the feature in linear bandit setting\\\hline
        $A_t$  & Action taken at round $t$ \\\hline
        $\|x\|_{M}$  & $\sqrt{x^\top M x}$ \\\hline
        $\A_t$  & Changing action set at round $t$ \\\hline
        $\tA_t
  = \{a \in \A_t:
  \max_{\btheta \in \mathcal{C}_t} \langle a, \btheta\rangle
  \geq \max_{a' \in \A_t}
  \min_{\btheta \in \mathcal{C}_t} \langle a', \btheta\rangle\}$  & Constraint action set at round $t$\\\hline
        $R_n$  & Cumulative regret till $n$\\\hline
        $V_t\coloneqq\sum_{s=1}^{t-1}A_sA^{\top}_{s}$  & Design matrix\\\hline
        $\btheta_{*}$  & Unknown model parameter\\\hline
        $\C_t$ & High-probability confidence set\\\hline
        $n$  & Total horizon\\\hline
        $Q_n
  = \sum_{t = 1}^n \max_{a \in \tA_t} \, (a^\top (\wtheta_t - \btheta_*))^2$  & Interpretability metric\\\hline
        $\Delta(a)\coloneqq \mu(a^*) - \mu(a)$  & Suboptimality gap of action $a$ \\\hline
        $T_t(a) \coloneqq \sum_{s = 1}^{t - 1} \I{A_s = a}$  & Number of times that action $a$ is taken up to round $t$\\\hline
        $\wmu_t(a) \coloneqq T_t^{-1}(a) \sum_{s = 1}^{t - 1} \I{A_s = a} Y_s$ & Empirical mean reward\\\hline
        $c_t(a)\coloneqq \sqrt{\frac{2\log(1/\delta)}{T_t(a)}}$  & Confidence width in $K$-armed bandits\\\hline
        $\sqrt{\beta_t(\delta)} = \sqrt{d \log((1 + t L^2 / \lambda) \delta)} + \lambda^{1 / 2} S$  & Confidence width in linear bandits\\\hline
        $\bI_d$  & $d$-dimensional identity matrix\\\hline
    \end{tabular}
    \vspace{1em}
    \caption{Table of Notations}
    \label{tab:my_label}
\end{table}






\newpage

\end{document}